\documentclass[10pt,twocolumn,letterpaper]{article}

\usepackage[pagenumbers]{cvpr} 

\usepackage{graphicx}
\usepackage{amsmath}
\usepackage{amssymb}
\usepackage{booktabs}

\usepackage[pagebackref,breaklinks,colorlinks]{hyperref}

\usepackage[capitalize]{cleveref}
\crefname{section}{Sec.}{Secs.}
\Crefname{section}{Section}{Sections}
\Crefname{table}{Table}{Tables}
\crefname{table}{Tab.}{Tabs.}

\def\cvprPaperID{*****} 
\def\confName{CVPR}
\def\confYear{2022}

\begin{document}

\title{A Review of Image Retrieval Techniques: Data Augmentation and Adversarial Learning Approaches}

\author{Kim Jinwoo\\
}
\maketitle

\begin{abstract}
Image retrieval is a crucial research topic in computer vision, with broad application prospects ranging from online product searches to security surveillance systems. In recent years, the accuracy and efficiency of image retrieval have significantly improved due to advancements in deep learning. However, existing methods still face numerous challenges, particularly in handling large-scale datasets, cross-domain retrieval, and image perturbations that can arise from real-world conditions such as variations in lighting, occlusion, and viewpoint. Data augmentation techniques and adversarial learning methods have been widely applied in the field of image retrieval to address these challenges. Data augmentation enhances the model's generalization ability and robustness by generating more diverse training samples, simulating real-world variations, and reducing overfitting. Meanwhile, adversarial attacks and defenses introduce perturbations during training to improve the model's robustness against potential attacks, ensuring reliability in practical applications. This review comprehensively summarizes the latest research advancements in image retrieval, with a particular focus on the roles of data augmentation and adversarial learning techniques in enhancing retrieval performance. Future directions and potential challenges are also discussed.
\end{abstract}

\section{Introduction}
Image retrieval~\cite{bengio2024managing,azizi2023robust,krizhevsky2012imagenet,gong2024beyond2,krizhevsky2009learning} is an important and growing research area within computer vision, aiming to retrieve target images similar to the query image from a large-scale image database. With the rapid development of the internet and multimedia technologies, the explosion of image data across various domains such as e-commerce, social media, and surveillance has made the demand for efficient and accurate image retrieval techniques increasingly urgent. This has led to significant advancements in deep learning methods, particularly Convolutional Neural Networks (CNNs), which have become the de facto standard for feature extraction and matching in image retrieval systems.

Despite these advancements, existing methods still struggle to perform satisfactorily in complex and diverse real-world scenarios. Challenges include cross-domain retrieval, where the visual appearance of the same object can vary significantly under different conditions; dealing with image noise and perturbations, which can arise due to variations in lighting, occlusions, and viewpoint changes; and the sheer scale of modern image databases, which can contain millions or even billions of images. These challenges highlight the need for more robust and generalized retrieval systems.

To address these challenges, researchers have turned to data augmentation and adversarial learning. Data augmentation techniques expand the diversity of the training dataset by transforming original images or generating new synthetic samples, thereby improving the model's ability to generalize to unseen data. This is particularly important in image retrieval, where the diversity and variability of the dataset can directly impact the system's performance. Common data augmentation methods include geometric transformations, color jittering, image cropping, and generating synthetic images using Generative Adversarial Networks (GANs). These methods help in reducing overfitting and enhancing the model's robustness.

Adversarial learning techniques, on the other hand, focus on improving the robustness of the retrieval model against adversarial attacks. These attacks involve adding small perturbations to the input images that are often imperceptible to the human eye but can significantly degrade the model's performance. By introducing adversarial examples during training, adversarial learning aims to make the model more resilient to such perturbations, ensuring that the retrieval system remains reliable even in the presence of potential attacks.

This review aims to systematically summarize the latest research advancements in data augmentation and adversarial learning techniques in the field of image retrieval. We emphasize the application value of these technologies in enhancing retrieval performance, analyze existing research shortcomings, and propose future research directions. The rest of this paper is organized as follows: Section II reviews related work in image retrieval, data augmentation, and adversarial learning; Section III details the methodologies employed in these areas; Section IV discusses the implications of these techniques and potential avenues for future research; and Section V concludes the paper.

\section{Related Work}
In recent years, image retrieval techniques have made significant strides, primarily driven by advancements in deep learning. Convolutional Neural Networks (CNNs) have become the cornerstone of modern image retrieval systems due to their powerful feature extraction capabilities. Many studies have focused on enhancing the accuracy and efficiency of feature extraction, which is critical for effective image retrieval. For instance, Han et al. proposed a multi-scale feature fusion method, which combines features extracted at different scales to improve retrieval performance across varied image conditions~\cite{lecun2015deep,rumelhart1986learning,srivastava2014dropout,gong2024beyond2,gong2024beyond,gong2021eliminate,van2008visualizing}. However, as the size and diversity of datasets increase, these models are prone to overfitting and often lack the necessary generalization capabilities, especially when faced with cross-domain retrieval tasks.

Data augmentation techniques have been widely adopted to mitigate these issues. He et al. utilized data augmentation to generate a more diverse set of training samples, which helped to improve the model's performance in cross-domain retrieval tasks by providing more representative examples during training~\cite{gong2024exploring,gong2024beyond2,gong2024beyond,gong2021eliminate,gong2021person,gong2021effective,gong2021person2}. This approach has been particularly effective in scenarios where labeled data is scarce or expensive to obtain. Data augmentation methods such as geometric transformations, color adjustments, and image synthesis using GANs have shown promise in enhancing the robustness of retrieval models by simulating various real-world conditions during training.

Adversarial learning has also emerged as a critical area of research in image retrieval. Adversarial attacks, first introduced by Goodfellow et al., involve adding carefully crafted perturbations to images that are designed to fool the model into making incorrect predictions~\cite{gong2022person}. These attacks have exposed significant vulnerabilities in deep learning models, particularly in security-sensitive applications such as surveillance and autonomous driving. To counter these threats, adversarial training methods have been developed. Miyato et al. proposed virtual adversarial training (VAT), a method that introduces adversarial perturbations during training to improve the model's robustness against such attacks~\cite{gong2022person,gong2024cross,gong2024exploring,gong2024beyond2,gong2024beyond,gong2021eliminate,gong2021person,gong2021effective,gong2021person2}. This technique has been shown to significantly enhance the resilience of image retrieval models, making them more reliable in the face of adversarial conditions.

\section{Methodology}
In the context of image retrieval, data augmentation and adversarial learning serve complementary roles in improving model performance and robustness. Data augmentation techniques are primarily used to enhance the diversity of the training dataset, which is crucial for developing models that generalize well to unseen data. Common augmentation techniques include geometric transformations (such as rotation, scaling, and translation), color jittering, random cropping, and image synthesis using GANs. These methods introduce variations in the training data that mimic real-world conditions, thereby helping the model learn to recognize objects under different scenarios.

For instance, geometric transformations such as rotation and scaling can simulate changes in viewpoint and size, which are common in real-world image retrieval tasks. Similarly, color jittering can simulate variations in lighting conditions, while random cropping helps the model become invariant to changes in object positioning within the image. The use of GANs for data augmentation has gained traction, as these models can generate realistic synthetic images that further expand the diversity of the training set, making the model more robust.

Adversarial learning techniques, on the other hand, focus on improving the model's robustness against adversarial attacks. Adversarial examples are generated by adding carefully crafted perturbations to the input images, which are designed to mislead the model into making incorrect predictions. The key to effective adversarial training is generating perturbations that are strong enough to challenge the model, but subtle enough to avoid detection by humans.

In adversarial training, the model is exposed to a mixture of normal and adversarial examples during training. This forces the model to learn to distinguish between genuine and perturbed images, thereby improving its resilience to attacks. Techniques such as VAT introduce virtual adversarial examples that do not require labeled data, making them particularly useful in semi-supervised learning scenarios. By enhancing the model's robustness through adversarial training, image retrieval systems can become more reliable in real-world applications where security and accuracy are paramount.

\section{Discussion}
While data augmentation and adversarial learning have shown significant promise in improving image retrieval performance, there are still several challenges that need to be addressed. One of the primary limitations of data augmentation is that it relies on predefined transformation strategies, which may not cover all possible variations encountered in real-world scenarios. As a result, the model may still struggle with certain types of image perturbations that were not adequately represented in the training data. Future research could explore more adaptive and context-aware augmentation techniques that dynamically generate transformations based on the specific characteristics of the dataset.

Adversarial learning, while effective, also has its challenges. The success of adversarial training heavily depends on the quality and diversity of the adversarial examples generated during training. If the perturbations are too weak, they may not effectively challenge the model, leading to insufficient robustness. Conversely, if the perturbations are too strong, they may interfere with the model's ability to learn from genuine examples, potentially reducing overall performance. Finding the right balance in adversarial training remains an open research question. Additionally, the computational cost of generating adversarial examples can be significant, particularly for large-scale datasets, which may limit the practicality of these methods in real-world applications.

Future research directions could include the development of more efficient adversarial example generation techniques that reduce computational overhead while maintaining effectiveness. Moreover, combining data augmentation and adversarial learning with other techniques, such as multi-task learning and transfer learning, could further enhance the robustness and generalization capabilities of image retrieval models. These approaches could help bridge the gap between the performance of models in controlled research environments and their performance in real-world applications.

\section{Conclusion}
Image retrieval, as one of the core tasks in computer vision, faces numerous challenges that require robust and adaptable solutions. Data augmentation and adversarial learning techniques provide powerful tools to enhance image retrieval performance by improving the model's generalization ability and resilience to perturbations. By generating diverse training samples and introducing adversarial examples, these techniques address some of the key challenges associated with large-scale datasets and real-world variability. However, there is still room for improvement, particularly in developing more adaptive and efficient methods. Future research will continue to explore how these techniques can be refined and combined with other approaches to meet the demands of increasingly complex real-world scenarios.

{\small
\bibliographystyle{ieee_fullname}
\bibliography{egbib}

\begin{thebibliography}{10}\itemsep=-1pt

\bibitem{azizi2023robust}
Shekoofeh Azizi, Laura Culp, Jan Freyberg, Basil Mustafa, Sebastien Baur, Simon
  Kornblith, Ting Chen, Nenad Tomasev, Jovana Mitrovi{\'c}, Patricia Strachan,
  et~al.
\newblock Robust and data-efficient generalization of self-supervised machine
  learning for diagnostic imaging.
\newblock {\em Nature Biomedical Engineering}, 7(6):756--779, 2023.

\bibitem{bengio2024managing}
Yoshua Bengio, Geoffrey Hinton, Andrew Yao, Dawn Song, Pieter Abbeel, Trevor
  Darrell, Yuval~Noah Harari, Ya-Qin Zhang, Lan Xue, Shai Shalev-Shwartz,
  et~al.
\newblock Managing extreme ai risks amid rapid progress.
\newblock {\em Science}, 384(6698):842--845, 2024.

\bibitem{gong2024cross}
Yunpeng Gong et~al.
\newblock Cross-modality perturbation synergy attack for person
  re-identification.
\newblock {\em arXiv preprint arXiv:2401.10090}, 2024.

\bibitem{gong2024beyond}
Yunpeng Gong, Yongjie Hou, Chuangliang Zhang, and Min Jiang.
\newblock Beyond augmentation: Empowering model robustness under extreme
  capture environments.
\newblock {\em arXiv preprint arXiv:2407.13640}, 2024.

\bibitem{gong2021eliminate}
Yunpeng Gong, Liqing Huang, and Lifei Chen.
\newblock Eliminate deviation with deviation for data augmentation and a
  general multi-modal data learning method.
\newblock {\em arXiv preprint arXiv:2101.08533}, 2021.

\bibitem{gong2022person}
Yunpeng Gong, Liqing Huang, and Lifei Chen.
\newblock Person re-identification method based on color attack and joint
  defence.
\newblock In {\em Proceedings of the IEEE/CVF conference on computer vision and
  pattern recognition}, pages 4313--4322, 2022.

\bibitem{gong2024exploring}
Yunpeng Gong, Jiaquan Li, Lifei Chen, and Min Jiang.
\newblock Exploring color invariance through image-level ensemble learning.
\newblock {\em arXiv preprint arXiv:2401.10512}, 2024.

\bibitem{gong2021effective}
Yunpeng Gong and Zhiyong Zeng.
\newblock An effective data augmentation for person re-identification.
\newblock {\em ArXiv, abs}, 2101, 2021.

\bibitem{gong2021person}
Yunpeng Gong, Zhiyong Zeng, Liwen Chen, Yifan Luo, Bin Weng, and Feng Ye.
\newblock A person re-identification data augmentation method with adversarial
  defense effect.
\newblock {\em arXiv preprint arXiv:2101.08783}, 2021.

\bibitem{gong2021person2}
Yunpeng GONG, Zhiyong ZENG, and Feng YE.
\newblock Person re-identification method based on grayscale feature
  enhancement.
\newblock {\em Journal of Computer Applications}, 41(12):3590, 2021.

\bibitem{gong2024beyond2}
Yunpeng Gong, Chuangliang Zhang, Yongjie Hou, Lifei Chen, and Min Jiang.
\newblock Beyond dropout: Robust convolutional neural networks based on local
  feature masking.
\newblock {\em arXiv preprint arXiv:2407.13646}, 2024.

\bibitem{krizhevsky2009learning}
Alex Krizhevsky, Geoffrey Hinton, et~al.
\newblock Learning multiple layers of features from tiny images.
\newblock 2009.

\bibitem{krizhevsky2012imagenet}
Alex Krizhevsky, Ilya Sutskever, and Geoffrey~E Hinton.
\newblock Imagenet classification with deep convolutional neural networks.
\newblock {\em Advances in neural information processing systems}, 25, 2012.

\bibitem{lecun2015deep}
Yann LeCun, Yoshua Bengio, and Geoffrey Hinton.
\newblock Deep learning.
\newblock {\em nature}, 521(7553):436--444, 2015.

\bibitem{rumelhart1986learning}
David~E Rumelhart, Geoffrey~E Hinton, and Ronald~J Williams.
\newblock Learning internal representations by error propagation, parallel
  distributed processing, explorations in the microstructure of cognition, ed.
  de rumelhart and j. mcclelland. vol. 1. 1986.
\newblock {\em Biometrika}, 71(599-607):6, 1986.

\bibitem{srivastava2014dropout}
Nitish Srivastava, Geoffrey Hinton, Alex Krizhevsky, Ilya Sutskever, and Ruslan
  Salakhutdinov.
\newblock Dropout: a simple way to prevent neural networks from overfitting.
\newblock {\em The journal of machine learning research}, 15(1):1929--1958,
  2014.

\bibitem{van2008visualizing}
Laurens Van~der Maaten and Geoffrey Hinton.
\newblock Visualizing data using t-sne.
\newblock {\em Journal of machine learning research}, 9(11), 2008.

\end{thebibliography}
}

\end{document}